\documentclass[
]{ceurart}

\usepackage{hyperref}

\newcommand{\fm}{FMML\textsuperscript{x }}
\newcommand{\xm}{XModeler\textsuperscript{ML }}
\newcommand{\mx}{UML-MX\textsuperscript{\textcopyright}}

\newcommand{\type}[1]{\texttt{#1}}
\newcommand{\class}[1]{\type{#1}}

\usepackage{listings}
\usepackage{csquotes}
\usepackage{url}

\begin{document}

\copyrightyear{2025}
\copyrightclause{Copyright for this paper by its authors.
  Use permitted under Creative Commons License Attribution 4.0
  International (CC BY 4.0).}

\conference{}

\title{How an unintended Side Effect of a Research Project led to Boosting the Power of UML}


\author[1]{Ulrich Frank}[%
orcid=0000-0002-8057-1836,
email=ulrich.frank@uni-due.de,
]
\cormark[1]
\address[1]{University of Duisburg-Essen, 45141 Essen, Germany}

\author[1]{Pierre Maier}[%
orcid=0009-0000-4594-6578,
email=pierre.maier@due.de,
]

\cortext[1]{Corresponding author.}

\begin{abstract}
This paper describes the design, implementation and use of a new UML modeling tool that represents a significant advance over conventional tools. Among other things, it allows the integration of class diagrams and object diagrams as well as the execution of objects. This not only enables new software architectures characterized by the integration of software with corresponding object models, but is also ideal for use in teaching, as it provides students with a particularly stimulating learning experience. A special feature of the project is that it has emerged from a long-standing international research project, which is aimed at a comprehensive multi-level architecture. The project is therefore an example of how research can lead to valuable results that arise as a side effect of other work.
\end{abstract}


\maketitle

\section{Introduction}

The history of science and engineering includes various examples of research projects where unintended side-effects led to new insights and, sometimes, groundbreaking inventions. To mention just a few well-known examples: In 1895, Wilhelm Conrad Röntgen examined electrical charges in a cathode tube, an almost airless glass tube. In the course of his experiments, his hand accidentally came between the cathode tube and a fluorescent screen, on which Röntgen was surprised to see the outline of the bones in his hand. In this way, he discovered a previously unknown form of radiation that could be used to visualize the human skeleton \cite{ARD}. A further example is Chemist Roy Plunkett's work on refrigerants that -- as a side-effect -- led to the invention of Teflon in 1938 \cite{APS}. While working on compact magnetic field tubes, Percy Spencer discovered by chance that food was heated in the vicinity of these tubes. This discovery inspired him to develop the microwave oven in 1945 \cite{Percy}.

While the story of the side-effect we describe in this paper is far from being as spectacular as the above examples, it is nevertheless suited to illustrate that research on languages and tools may produce side-effects of remarkable benefit. The research project from which this by-product emerged, \emph{Language Engineering for Multi-Level Modeling} (LE4MM), is unusual in that it ran over a long period of more than 15 years \citep{Frank.2023}. It is aimed at the ambitious goal of developing a comprehensive environment that enables, among other things, a common representation of programs, models and corresponding languages. Looking back, it seems almost ironic that the original research project was motivated not least by the desire to eliminate the limitations of MOF. So we were by no means interested in promoting the use of UML. Instead, we wanted to make it redundant. The unintended offspring of the project LE4MM is a UML modeling tool that literally takes the creation and use of class and object diagrams to a new level. It also provides effective support for teaching object-oriented modeling because it offers students an unprecedented learning experience. 


\section{Primary Focus: Multi-Level Language Architectures}
The motivation for the project LE4MM goes back to the nineties of last century. In that time, our research group focussed on the development of domain-specific modeling languages (DSMLs) and corresponding modeling tools. 
This work led to the specification of many DSMLs, especially in the context of enterprise modeling. Despite their obvious advantages, the work on these DSMLs was characterized by increasing frustration. This frustration was mainly caused by principal limitations of modeling and, more important, of programming languages.

Among other things, these languages are restricted to the dichotomy of classes and objects, where classes cannot have state. Therefore, classes specified, e.g., in a UML class diagram have to be implemented as objects in a corresponding modeling tool. As a consequence, they cannot be instantiated within the tool -- even though that should be possible with respect to their conceptual status. Hence, code needs to be generated from models, resulting in two separate representations, the synchronization of which is hardly possible in the long run. For the same reason, it is not possible to integrate DSMLs and corresponding models in the same tool. In addition, the expressiveness of DSMLs is restricted in the sense that it is not possible to represent all relevant domain knowledge within a DSML, which leads to redundant specifications with the models created with these DSMLs. For a comprehensive analysis of these limitations see \cite{Frank.2022}.

Multi-level modeling is based on a different perspective on object-oriented modeling and corresponding modeling languages. First, it allows for an arbitrary number of classification levels. Second, every class, no matter at what level, is an object that can have state and execute operations. Finally, classes at level n may constrain state and behavior not only of those objects that are direct instances, but also of objects at levels lower than n-1. This allows for a natural approach to create languages and models. A DSML does not have to be specified with a generic meta-modeling language, but may be defined with a less specific DSML. That corresponds directly to the evolution of technical languages, which are usually defined using a previously introduced, more general technical language. In addition, the language engineering environment \xm features a common representation of models and programs. Therefore, it is not required to generate code from models. Instead models serve as an additional representation that can be navigated at run time during the entire lifetime of a software system.

\begin{figure}[htb]
\begin{center}
\includegraphics[width=0.8\textwidth]{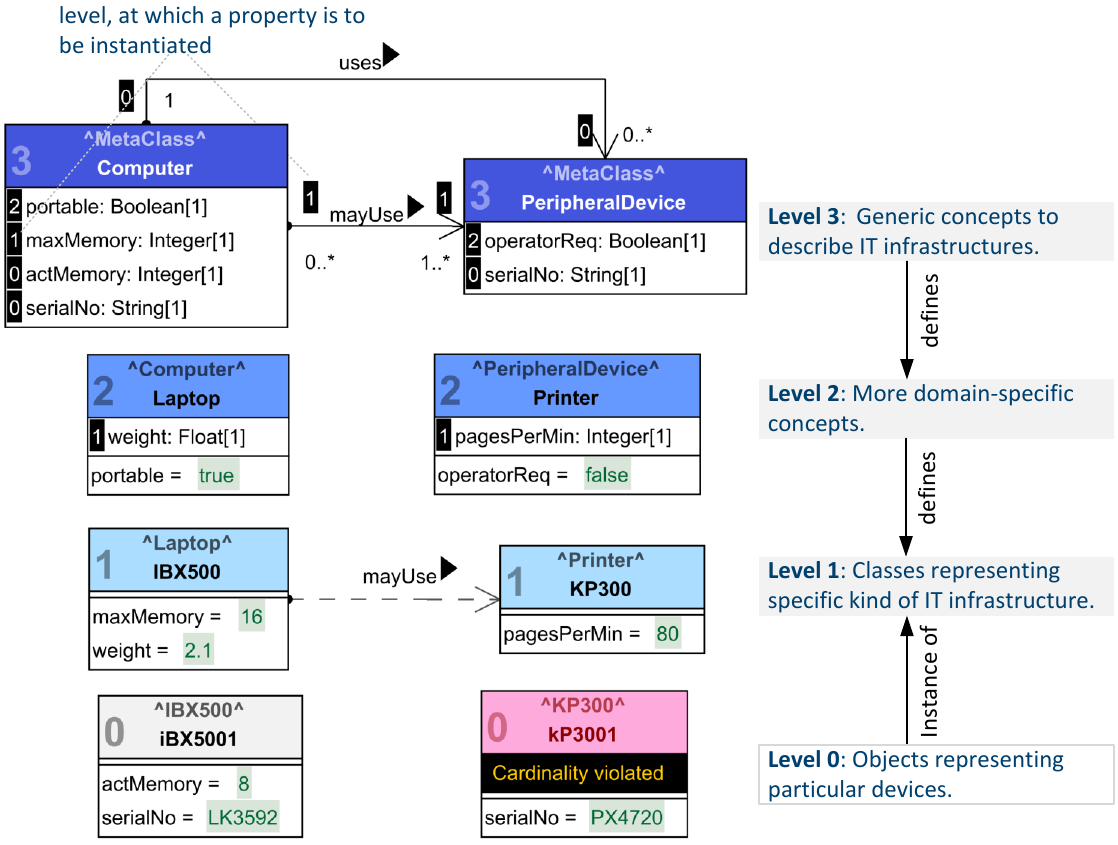}
\caption{Illustration of a multi-level hierarchy with \fm}
\label{figure:mlm}
\end{center}
\end{figure}

The main results that have been achieved so far comprise the \fm (Flexible and executable Multi-Level Language) \cite{Frank.2014e}, the \xm \cite{Frank.2022b}, a comprehensive language engineering, modeling and execution environment, and an accompanying design method \cite{Frank.2021d}. The \fm is specified through an extension of XCore, the meta model of the previously developed XModeler \cite[p. 40]{Clark.2008b}. The web pages of the LE4MM project (\url{www.le4mm.org}) provide various examples and screencasts that demonstrate the benefits of multi-level modeling. Fig. \ref{figure:mlm} illustrates the use of a multi-level language hierarchy with an example created with the \fm. Level 3 represents a general language for modeling IT infrastructures. It includes all invariant knowledge available about the domain, including, e.g., that a particular peripheral device has a serial number, or that it must be connected to a computer. In the example the latter constraint is violated. The concepts defined on L3 are used to define the language on L2, which in turn serves the specification of concepts at L1.

\section{Discovery of the Side-Effect: Not by Chance}

The advantages offered by multi-level language architectures are substantial and undeniable. They promote reuse and the reduction of redundancy. At the same time, they are suited to increase developers' productivity, since the specification of DSMLs and/or models does not require anymore to start from scratch. Also, and this is a compelling reason for the superiority of multi-level architectures, they allow for everything that can be done with traditional architectures and, in addition, for much more. Nevertheless, multi-level modeling, which has evolved over a period of more than twenty years \cite{Atkinson.Kuhne_2001}, has never made it to the mainstream, but stayed mainly within the niche of an enthusiastic, but small community.

Since it was not enough to point out the  -- from our perspective -- obvious advantages of multi-level architectures, we thought about creating additional incentives for trying out the new paradigm. A number of corresponding approaches are discussed in \cite{Frank.2022}. Among them was a strategy called \enquote{intrusion through the backdoor,} which aimed at offering users of UML a clearly improved experience -- both for professional and novice users -- to finally raise their interest in multi-level modeling.

\emph{Focus on Experienced Users of UML.} All UML modeling tools we know of suffer from two related shortcomings. First, they require separate representations of classes in class diagrams and objects in object diagrams. Modelers need to keep the two representations in sync. Automated consistency checks, if available at all, are severely limited. They often have difficulties accounting for, inter alia, data types, multiplicities, and changing class names. Separate representations are a consequence of the fact that classes are represented as objects on M0 in traditional UML modeling tools. Hence, they simply do not allow for actual instantiation. Due to this restriction, objects represented in a UML modeling editor can also not be executed. Instead, the approach of choice would be to generate code from an object model, which results in the often painful need for synchronizing model and code. This leads to two \enquote{still frequently encountered but equally unsatisfactory} options, as Meyer \cite[pp. 22-23]{Meyer_1997} notes, where developers are forced to use a language suited exclusively either for design (e.g., UML) or for implementation (e.g., Java). While many UML users potentially suffer from these shortcomings, they probably accept them as unavoidable. 

A modeling tool that offers UML, but features the integration of class and object diagrams -- and allows for directly executing objects should represent a convincing incentive to substitute the tools used so far.

\emph{Focus on Teaching.} When we discussed incentives for UML users to switch to a UML editor based on a multi-level architectures, we immediately realized that it was also suited to help with a challenge that had concerned us for some time. Students in the Bachelor program struggled with object-oriented modeling, even though the corresponding course had been thoroughly designed and improved over many years. It turned out that students often had problems with abstraction. In particular, they struggled with the distinction of objects and classes. A UML modeling tool that demonstrates the interplay of classes and objects should support them with developing an appropriate understanding of both. Furthermore, that is our hope, it could prepare students for appreciating the benefit of additional abstractions as they are enabled by multi-level language architectures.

\section{Turning a Side-Effect into a Research Result}
The idea to develop a powerful UML editor as an offspring of the \xm was so tempting not least because the related effort seemed to be manageable. At first, it required adapting the meta model underlying the \fm. In addition, the concrete syntax of the \fm and the user interface of the diagram editor had to be adapted accordingly. During the work on the UML editor, which we later called \mx, we decided to enhance the UML with a few features that are useful without preventing regular use of UML. In particular, these add-ons comprise constraints that are specified with XOCL (eXecutable Object Constraint Language), an extended, executable version of OCL \cite{TonyClark.2018}. They are checked by the tool immediately after they were specified. Furthermore, we extended UML with delegation including corresponding execution semantics.

\subsection{Foundational Metamodel and Modifications of the \xm}
\label{sec:metamodel}
The UML can be regarded as a subset of an object-oriented multi-level modeling language like the \fm. Hence, adapting the meta model of the \fm \cite{Frank2018,Frank&Topel2024} recommends fading out all specific multi-level features, such as classes at levels above 1 and intrinsic properties. 
We refer to this adapted version of the \fm, a monotonic extension of the UML, as UML++ \cite{Maier.2024}. While the \fm allows classes to be at any level, UML++ restricts classes to be at level 1 only. Per default, \class{MetaClass} is at level 2. Hence, UML++ classes that are instantiated from \class{MetaClass} are at level 1. 
The important difference to traditional UML editors is that classes are implemented at level 1, not as objects at 0. Therefore, they can be instantiated into objects at level 0, and reside in the same namespace as these. Also, objects can execute the operations specified in their classes. Consequently, and different from the \fm, properties such as attributes, operations and associations have to be instantiated always at level 0. The discrepancy between analysis and implementation languages criticized by Meyer has thus been largely overcome.

Features of the \xm that that are irrelevant to modeling with UML were hidden from the user. We simplified most add/change dialogues and stripped them of any multi-level-modeling capabilities. Also, the concrete syntax editor that enables the development of custom notations for DSMLs was faded out. Currently, \mx supports class and object diagrams only. \mx is described in more detail in \cite{Frank&Maier2025}.



\subsection{A new Way to Model Class and Object Diagrams}
\begin{figure}[htb]
\begin{center}
\includegraphics[width=.9\textwidth]{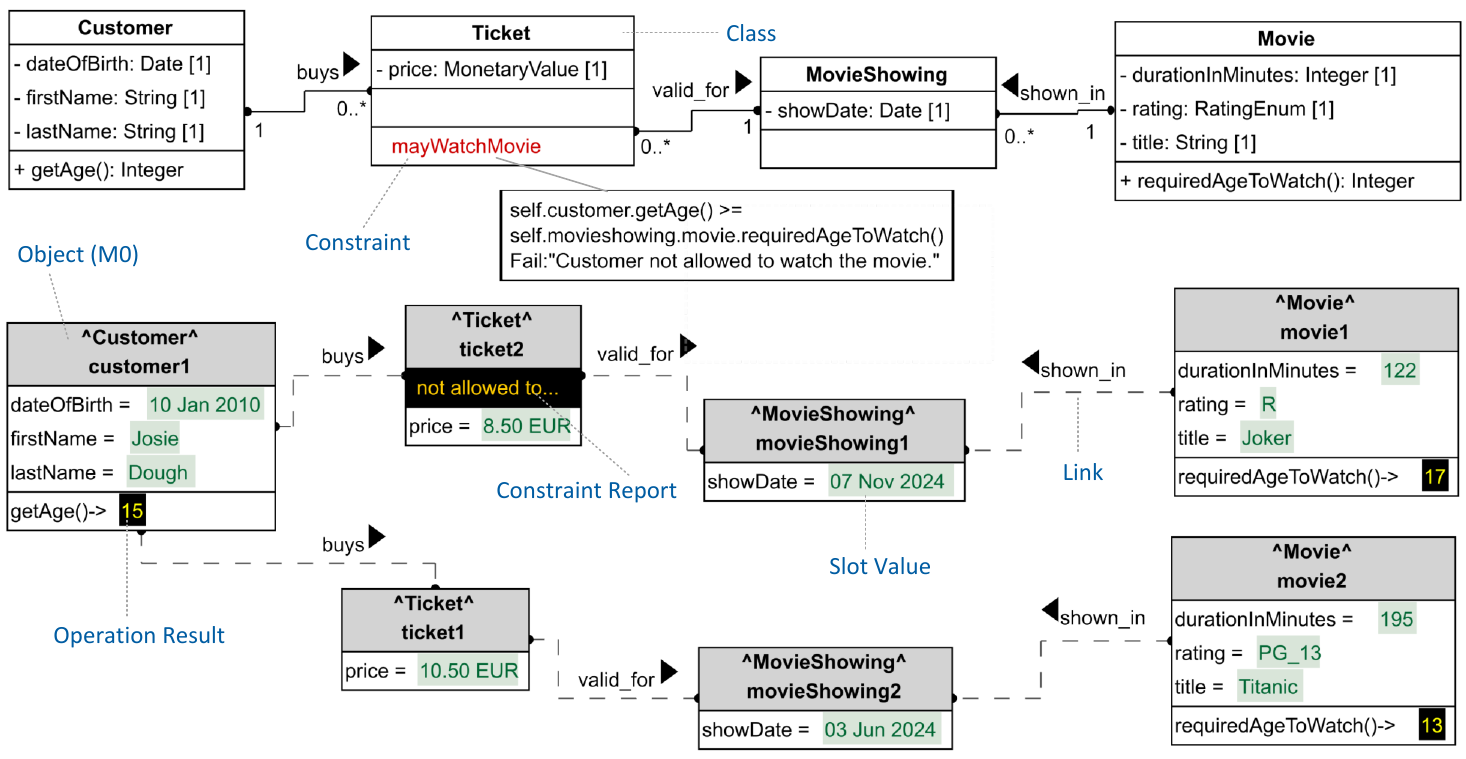}
\caption{Example UML++ diagram for modeling a cinema}
\label{figure:uml-example}
\end{center}
\end{figure}

An example UML++ diagram is displayed in Fig. \ref{figure:uml-example}. Different from regular UML models, classes and objects can be, but do not have to be, modeled within the same diagram. Classes include an additional compartment if constraints are present, as in the example class \class{Ticket}. Upon the definition of a new class, it appears as an entry in the palette (see screenshot in Fig. \ref{figure:del-example}), which can be dragged to the diagram to instantiate an object. All constraints are evaluated at runtime and custom error messages are shown whenever a constraint is violated. A corresponding error message can be partly seen in the object \class{ticket2}. Within the tool, error messages are shown in the style of running LED banners.

The \fm notation was adapted to that of UML. Levels of classes and objects, as well as the instantiation levels of class properties, are faded out. To support a quick visual distinction between objects and classes, the compartment for object names has a gray background color. 
With UML++, slot values of objects can either be entered by users or computed. Also, users may implement operations for classes. Custom operations and constraints are written in XOCL (see example in Fig. \ref{figure:del-example}). These operations are executed by instances of the corresponding class as soon as they are triggered -- either by user request or upon a change of object state. Corresponding return values are shown in each object, if a certain flag is set with the implementation of the operation. 

All attributes must be specified with a data type, a custom class or an enumeration. Data types need to be selected from the default \fm data types provided originally with the \xm. These include, next to common types such as \class{String} or \class{Date}, also \class{MonetaryValue} which can be used to model prices (see the attribute \class{price} in the class \class{Ticket}). 

\subsection{Specific Support for Teaching}
When we taught UML, we increasingly became aware that students struggled with discriminating objects from classes. The traditional UML tool we used back then forced students to develop two separate diagrams to model classes and corresponding object. That contributed to students having difficulties, e.g., with understanding the effect the modification of a class would have on its instances. The introduction of the \mx proved to address these difficulties successfully, even though it was not a silver bullet that turned every student quickly into an UML expert.



\begin{figure}[htb]
\begin{center}
\includegraphics[width=0.95\textwidth]{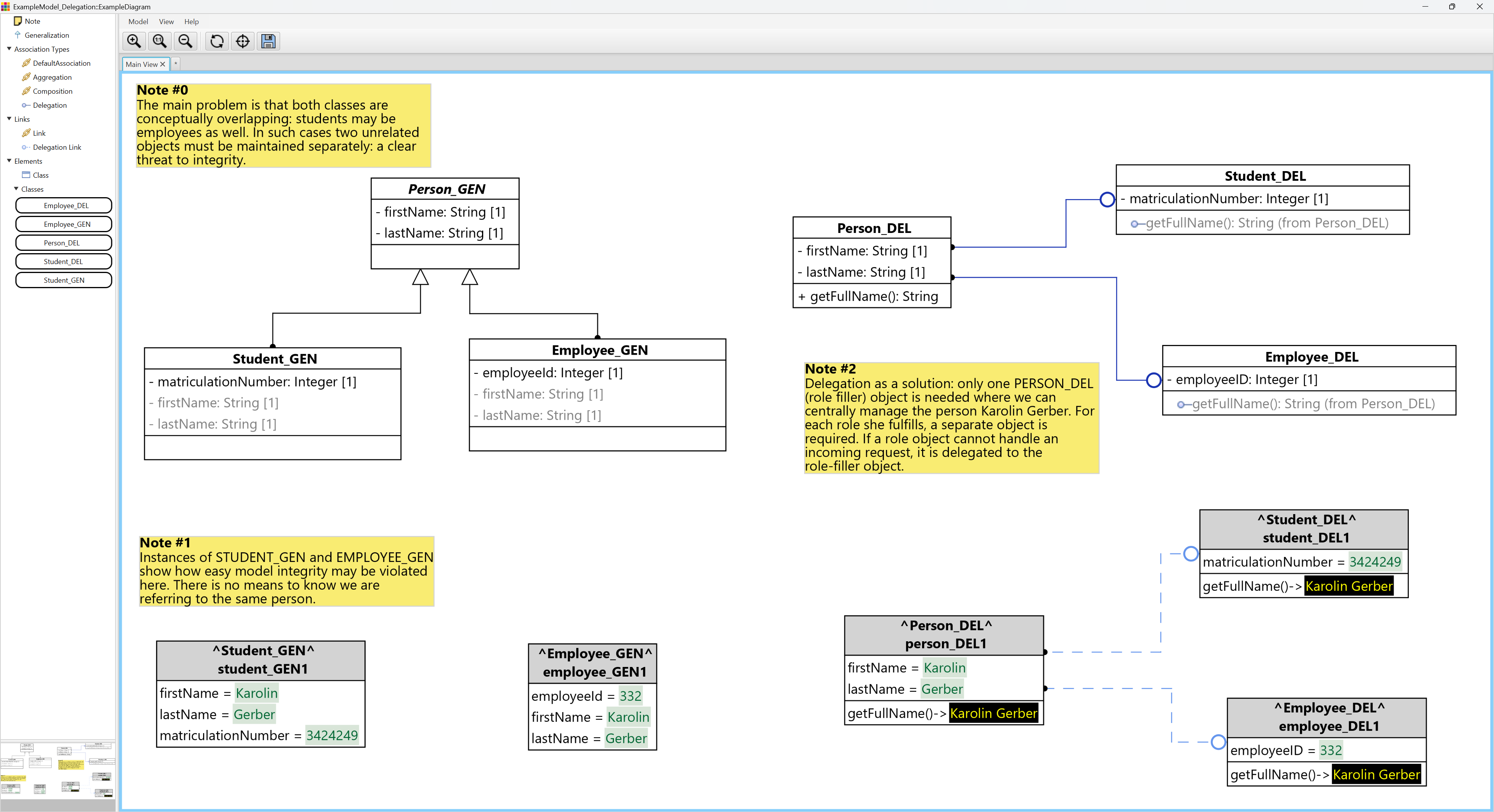}
\caption{Example model for the learning unit on delegation opened in \mx}
\label{figure:del-example}
\end{center}
\end{figure}

This pleasing result motivated us to expand the tool's support for students.  since they still experienced problems with distinguishing attributes from associated object, with using generalization/specialization properly, etc. Therefore, we focused on more individual support for learning UML, since this was lacking due to the number of participants being around 200. To provide individual support for learning UML, we developed ten learning units that cover topics ranging from teaching fundamental constructs such as objects and associations to more advanced object-modeling principles such as model circles and derivable attributes. Each of these learning units consists of a set of learning objectives, a theoretical background that includes fundamental explanations as well as an example model illustrating the core concepts of the learning unit, and a set of modeling exercises that can be completed within the tool. The example model for the learning unit \emph{Pitfalls of Specialization and Delegation} is shown in Fig. \ref{figure:del-example}. All learning units can be accessed for free in \mx. A comprehensive description of features especially designed for supporting students is provided in \cite{Maier.2024}. \mx can be downloaded at \url{www.le4mm.org/uml-mx}, where also further examples and screencasts are available. 


\section{Conclusions and Future Work}

We are convinced that the tool represents significant progress -- not only with regard to the creation and use of UML class and object diagrams but also for model-driven software development. Initial publications \cite{Maier.2024, Frank&Maier2025} and the provision of the tool together with numerous examples have led to considerable attention being paid to \mx, and we hope for a growing number of users, especially to support teaching. Nevertheless, it remains our aim to replace traditional language architectures like MOF as well as corresponding object-oriented modeling languages with significantly more efficient multi-level architectures. We hope that the tool will serve as a catalyst for this by inspiring users to finally overcome the limits of UML. In addition, the project illustrates that it may pay off to think beyond original project goals from time to time.


\bibliography{bib}

\end{document}